# MasakhaNER: Named Entity Recognition for African Languages

**David Ifeoluwa Adelani**[1*], **Jade Abbott**[2*], **Graham Neubig**[3], **Daniel D'souza**[4*],
**Julia Kreutzer**[5*], **Constantine Lignos**[6*], **Chester Palen-Michel**[6*], **Happy Buzaaba**[7*],
**Shruti Rijhwani**[3], **Sebastian Ruder**[8], **Stephen Mayhew**[9], **Israel Abebe Azime**[10*],
**Shamsuddeen H. Muhammad**[11,12*], **Chris Chinenye Emezue**[13*], **Joyce Nakatumba-Nabende**[14*],
**Perez Ogayo**[15*], **Anuoluwapo Aremu**[16*], **Catherine Gitau**[*], **Derguene Mbaye**[*],
**Jesujoba Alabi**[17*], **Seid Muhie Yimam**[18], **Tajuddeen Rabiu Gwadabe**[19*], **Ignatius Ezeani**[20*],
**Rubungo Andre Niyongabo**[21*], **Jonathan Mukiibi**[14], **Verrah Otiende**[22*],
**Iroro Orife**[23*], **Davis David**[*], **Samba Ngom**[*], **Tosin Adewumi**[24*],
**Paul Rayson**[20], **Mofetoluwa Adeyemi**[*], **Gerald Muriuki**[14], **Emmanuel Anebi**[*],
**Chiamaka Chukwuneke**[20], **Nkiruka Odu**[25], **Eric Peter Wairagala**[14],
**Samuel Oyerinde**[*], **Clemencia Siro**[*], **Tobius Saul Bateesa**[14], **Temilola Oloyede**[*],
**Yvonne Wambui**[*], **Victor Akinode**[*], **Deborah Nabagereka**[14], **Maurice Katusiime**[14],
**Ayodele Awokoya**[26*], **Mouhamadane MBOUP**[*], **Dibora Gebreyohannes**[*], **Henok Tilaye**[*],
**Kelechi Nwaike**[*], **Degaga Wolde**[*], **Abdoulaye Faye**[*], **Blessing Sibanda**[27*],
**Orevaoghene Ahia**[28*], **Bonaventure F. P. Dossou**[29*], **Kelechi Ogueji**[30*],
**Thierno Ibrahima DIOP**[*], **Abdoulaye Diallo**[*], **Adewale Akinfaderin**[*],
**Tendai Marengereke**[*], and **Salomey Osei**[10*]

[*] Masakhane NLP, [1] Spoken Language Systems Group (LSV), Saarland University, Germany
[2] Retro Rabbit, [3] Language Technologies Institute, Carnegie Mellon University
[4] ProQuest, [5] Google Research, [6] Brandeis University, [8] DeepMind, [9] Duolingo
[7] Graduate School of Systems and Information Engineering, University of Tsukuba, Japan.
[10] African Institute for Mathematical Sciences (AIMS-AMMI), [11] University of Porto
[12] Bayero University, Kano, [13] Technical University of Munich, Germany
[14] Makerere University, Kampala, Uganda, [15] African Leadership University, Rwanda
[16] University of Lagos, Nigeria, [17] Max Planck Institute for Informatics, Germany.
[18] LT Group, Universität Hamburg, [19] University of Chinese Academy of Science, China
[20] Lancaster University, [21] University of Electronic Science and Technology of China, China.
[22] United States International University - Africa (USIU-A), Kenya. [23] Niger-Volta LTI
[24] Luleå University of Technology, [25] African University of Science and Technology, Abuja
[26] University of Ibadan, Nigeria, [27] Namibia University of Science and Technology
[28] Instadeep, [29] Jacobs University Bremen, Germany, [30] University of Waterloo

## Abstract

We take a step towards addressing the under-representation of the African continent in NLP research by bringing together different stakeholders to create the first large, publicly available, high-quality dataset for named entity recognition (NER) in ten African languages. We detail the characteristics of these languages to help researchers and practitioners better understand the challenges they pose for NER tasks. We analyze our datasets and conduct an extensive empirical evaluation of state-of-the-art methods across both supervised and transfer learning settings. Finally, we release the data, code, and models to inspire future research on African NLP [1].

## 1 Introduction

Africa has over 2,000 spoken languages (Eberhard et al., 2020); however, these languages are scarcely represented in existing natural language processing (NLP) datasets, research, and tools (Martinus and Abbott, 2019). ∀ et al. (2020) investigate the reasons for these disparities by examining how NLP for low-resource languages is constrained by several societal factors. One of these factors is the geographical and language diversity of NLP researchers. For example, of the 2695 affiliations of authors whose works were published at the five major NLP conferences in 2019, only five were from African institutions (Caines, 2019). Conversely, many NLP tasks such as machine translation, text classification, part-of-speech tagging, and named entity recognition would benefit from the knowledge of native speakers who are involved in the development of datasets and models.

---
[1] https://git.io/masakhane-ner

In this work, we focus on named entity recognition (NER)—one of the most impactful tasks in NLP (Sang and De Meulder, 2003; Lample et al., 2016). NER is an important information extraction task and an essential component of numerous products including spell-checkers, localization of voice and dialogue systems, and conversational agents. It also enables identifying African names, places and organizations for information retrieval. African languages are under-represented in this crucial task due to lack of datasets, reproducible results, and researchers who understand the challenges that such languages present for NER.

In this paper, we take an initial step towards improving representation for African languages for the NER task, making the following contributions:

(i) We bring together language speakers, dataset curators, NLP practitioners, and evaluation experts to address the challenges facing NER for African languages. Based on the availability of online news corpora and language annotators, we develop NER datasets, models, and evaluation covering ten widely spoken African languages.

(ii) We curate NER datasets from local sources to ensure relevance of future research for native speakers of the respective languages.

(iii) We train and evaluate multiple NER models for all ten languages. Our experiments provide insights into the transfer across languages, and highlight open challenges.

(iv) We release the datasets, code, and models to facilitate future research on the specific challenges raised by NER for African languages.

## 2 Related Work

**African NER datasets** NER is a well-studied sequence labeling task (Yadav and Bethard, 2018) and has been the subject of many shared tasks in different languages (Tjong Kim Sang, 2002; Tjong Kim Sang and De Meulder, 2003; Sangal et al., 2008; Shaalan, 2014; Benikova et al., 2014). However, most of the available datasets are in high-resource languages. Although there have been efforts to create NER datasets for lower-resourced languages, such as the WikiAnn corpus (Pan et al., 2017) covering 282 languages, such datasets consist of "silver-standard" labels created by transferring annotations from English to other languages through cross-lingual links in knowledge bases. Because the WikiAnn corpus data comes from Wikipedia, it includes some African languages; though most have fewer than 10k tokens.

Other NER datasets for African languages include SADiLaR (Eiselen, 2016) for ten South African languages based on government data, and small corpora of fewer than 2K sentences for Yorùbá (Alabi et al., 2020) and Hausa (Hedderich et al., 2020). Additionally, the LORELEI language packs (Strassel and Tracey, 2016) include some African languages (Yorùbá, Hausa, Amharic, Somali, Twi, Swahili, Wolof, Kinyarwanda, and Zulu), but are not publicly available.

**NER models** Popular sequence labeling models for NER include the CRF (Lafferty et al., 2001), CNN-BiLSTM (Chiu and Nichols, 2016), BiLSTM-CRF (Huang et al., 2015), and CNN-BiLSTM-CRF (Ma and Hovy, 2016). The traditional CRF makes use of hand-crafted features like part-of-speech tags, context words and word capitalization. Neural NER models on the other hand are initialized with word embeddings like Word2Vec (Mikolov et al., 2013), GloVe (Pennington et al., 2014) and FastText (Bojanowski et al., 2017). More recently, pre-trained language models such as BERT (Devlin et al., 2019), RoBERTa (Liu et al., 2019), and LUKE (Yamada et al., 2020) have been applied to produce state-of-the-art results for the NER task. Multilingual variants of these models like mBERT and XLM-RoBERTa (Conneau et al., 2020) make it possible to train NER models for several languages using transfer learning. Language-specific parameters and adaptation to unlabeled data of the target language have yielded further gains (Pfeiffer et al., 2020a,b).

## 3 Focus Languages

Table 1 provides an overview of the languages considered in this work, their language family, number of speakers and the regions in Africa where they are spoken. We chose to focus on these languages due to the availability of online news corpora, annotators, and most importantly because they are widely spoken native African languages. Both region and language family might indicate a notion of proximity for NER, either because of linguistic features shared within that family, or because data sources cover a common set of locally relevant entities. We highlight language specifics for each language to illustrate the diversity of this selection of languages in Section 3.1, and then showcase the differences in named entities across these languages in Section 3.2.

| Language | Family | Speakers | Region |
|---|---|---|---|
| Amharic | Afro-Asiatic-Ethio-Semitic | 33M | East |
| Hausa | Afro-Asiatic-Chadic | 63M | West |
| Igbo | Niger-Congo-Volta-Niger | 27M | West |
| Kinyarwanda | Niger-Congo-Bantu | 12M | East |
| Luganda | Niger-Congo-Bantu | 7M | East |
| Luo | Nilo Saharan | 4M | East |
| Nigerian-Pidgin | English Creole | 75M | West |
| Swahili | Niger-Congo-Bantu | 98M | Central & East |
| Wolof | Niger-Congo-Senegambia | 5M | West & NW |
| Yorùbá | Niger-Congo-Volta-Niger | 42M | West |

Table 1: Language, family, number of speakers (Eberhard et al., 2020), and regions in Africa.

### 3.1 Language Characteristics

**Amharic** (amh) uses the Fidel script consisting of 33 basic scripts (ሀ (hä) ለ (lä) መ (mä) ሠ (šä) ...), each of them with at least 7 vowel sequences (such as ሀ (hä) ሁ (hu) ሂ (hī) ሃ (ha) ሄ (hē) ህ (hi) ሆ (ho)). This results in more than 231 characters or Fidels. Numbers and punctuation marks are also represented uniquely with specific Fidels (፩ (1), ፪ (2), ... and ። (.), !(!), ፤ (;),).

**Hausa** (hau) has 23-25 consonants, depending on the dialect and five short and five long vowels. Hausa has labialized phonemic consonants, as in /gw/ e.g. 'agwagwa.' As found in some African languages, implosive consonants also exist in Hausa, e.g. 'b, 'd, etc as in 'barna'. Similarly, the Hausa approximant 'r' is realized in two distinct manners: roll and trill, as in 'rai' and 'ra'ayi', respectively.

**Igbo** (ibo) is an agglutinative language, with many frequent suffixes and prefixes (Emenanjo, 1978). A single stem can yield many word-forms by addition of affixes that extend its original meaning (Onyenwe and Hepple, 2016). Igbo is also tonal, with two distinctive tones (high and low) and a down-stepped high tone in some cases. The alphabet consists of 28 consonants and 8 vowels (A, E, I, Ị, O, Ọ, U, Ụ). In addition to the Latin letters (except *c*), Igbo contains the following digraphs: (ch, gb, gh, gw, kp, kw, nw, ny, sh).

**Kinyarwanda** (kin) makes use of 24 Latin characters with 5 vowels similar to English and 19 consonants excluding q and x. Moreover, Kinyarwanda has 74 additional complex consonants (such as mb, mpw, and njyw). (Government, 2014) It is a tonal language with three tones: low (no diacritic), high (signaled by "/") and falling (signaled by "^"). The default word order is Subject-Verb-Object.

**Luganda** (lug) is a tonal language with subject-verb-object word order. The Luganda alphabet is composed of 24 letters that include 17 consonants (p, v, f, m, d, t, l, r, n, z, s, j, c, g), 5 vowel sounds represented in the five alphabetical symbols (a, e, i, o, u) and 2 semi-vowels (w, y). It also has a special consonant $ng'$.

**Luo** (luo) is a tonal language with 4 tones (high, low, falling, rising) although the tonality is not marked in orthography. It has 26 Latin consonants without Latin letters (c, q, v, x and z) and additional consonants (ch, dh, mb, nd, ng', ng, ny, nj, th, sh). There are nine vowels (a, e, i, o, u, ɛ, ɛ, ɔ, ʊ) which are distinguished primarily by advanced tongue root (ATR) harmony (De Pauw et al., 2007).

**Nigerian-Pidgin** (pcm) is a largely oral, national lingua franca with a distinct phonology from English, its lexifier language. Portuguese, French, and especially indigenous languages form the substrate of lexical, phonological, syntactic, and semantic influence on Nigerian-Pidgin (NP). English lexical items absorbed by NP are often phonologically closer to indigenous Nigerian languages, notably in the realization of vowels. As a rapidly evolving language, the NP orthography is undergoing codification and indigenization (Offiong Mensah, 2012; Onovbiona, 2012; Ojarikre, 2013).

**Swahili** (swa) is the most widely spoken language on the African continent. It has 30 letters including 24 Latin letters without characters (q and x) and six additional consonants (ch, dh, gh, ng', sh, th) unique to Swahili pronunciation.

**Wolof** (wol) has an alphabet similar to that of French. It consists of 29 characters, including all letters of the French alphabet except H, V and Z. It also includes the characters Ŋ ("ng", lowercase: ŋ) and Ñ ("gn" as in Spanish). Accents are present, but limited in number (À, É, Ë, Ó). However, unlike many other Niger-Congo languages, Wolof is not a tonal language.

| Language | Sentence |
|---|---|
| English | The Emir of Kano turbaned Zhang who has spent 18 years in Nigeria |
| Amharic | የካኖ ኢምር በናይጀርያ ፩፰ ዓመት ያሳለፈውን ዛንግን ዋና መሪ አደረጉት |
| Hausa | Sarkin Kano yayi wa Zhang wanda yayi shekara 18 a Najeriya sarauta |
| Igbo | Onye Emir nke Kano kpubere Zhang okpu onye nke nọgoro afọ iri na asatọ na Naijiria |
| Kinyarwanda | Emir w'i Kano yimitse Zhang wari umaze imyaka 18 muri Nijeriya |
| Luganda | Emir w'e Kano yatikkidde Zhang amaze emyaka 18 mu Nigeria |
| Luo | Emir mar Kano ne orwakone turban Zhang ma osedak Nigeria kwuom higni 18 |
| Nigerian-Pidgin | Emir of Kano turban Zhang wey don spend 18 years for Nigeria |
| Swahili | Emir wa Kano alimvisha kilemba Zhang ambaye alikaa miaka 18 nchini Nigeria |
| Wolof | Emiiru Kanó dafa kaala kii di Zhang mii def Nigeria fukki at ak juróom ñett |
| Yorùbá | Ẹ́mía ìlú Kánò wé láwàní lé orí Zhang ẹni tí ó ti lo ọdún méjìdínlógún ní orílẹ̀-èdè Nàìjíríà |

Table 2: Example of named entities in different languages. PER, LOC, and DATE are in colours purple, orange, and green respectively.

**Yorùbá** (yor) has 25 Latin letters without the Latin characters (c, q, v, x and z) and with additional letters (ẹ, gb, ṣ, ọ). Yorùbá is a tonal language with three tones: low ("\"), middle ("–", optional) and high ("/"). The tonal marks and underdots are referred to as diacritics and they are needed for the correct pronunciation of a word. Yorùbá is a highly isolating language and the sentence structure follows Subject-Verb-Object.

### 3.2 Named Entities

Most of the work on NER is centered around English, and it is unclear how well existing models can generalize to other languages in terms of sentence structure or surface forms. In Hu et al. (2020)'s evaluation on cross-lingual generalization for NER, only two African languages were considered and it was seen that transformer-based models particularly struggled to generalize to named entities in Swahili. To highlight the differences across our focus languages, Table 2 shows an English[2] example sentence, with color-coded PER, LOC, and DATE entities, and the corresponding translations. The following characteristics of the languages in our dataset could pose challenges for NER systems developed for English:

- Amharic shares no lexical overlap with the English source sentence.
- While "Zhang" is identical across all Latin-script languages, "Kano" features accents in Wolof and Yorùbá due to its localization.
- The Fidel script has no capitalization, which could hinder transfer from other languages.
- Igbo, Wolof, and Yorùbá all use diacritics, which are not present in the English alphabet.
- The surface form of named entities (NE) is the same in English and Nigerian-Pidgin, but there exist lexical differences (e.g. in terms of how time is realized).
- Between the 10 African languages, "Nigeria" is spelled in 6 different ways.
- Numerical "18": Igbo, Wolof and Yorùbá write out their numbers, resulting in different numbers of tokens for the entity span.

## 4 Data and Annotation Methodology

Our data was obtained from local news sources, in order to ensure relevance of the dataset for native speakers from those regions. The dataset was annotated using the ELISA tool (Lin et al., 2018) by native speakers who come from the same regions as the news sources and volunteered through the *Masakhane* community[3]. Annotators were not paid but are all part of the authors of this paper. The annotators were trained on how to perform NER annotation using the MUC-6 annotation guide[4]. We annotated four entity types: Personal name (PER), Location (LOC), Organization (ORG), and date & time (DATE). The annotated entities were inspired by the English CoNLL-2003 Corpus (Tjong Kim Sang, 2002). We replaced the MISC tag with the DATE tag following Alabi et al. (2020) as the MISC tag may be ill-defined and cause disagreement among non-expert annotators. We report the number of annotators as well as general statistics of the datasets in Table 3. For each language, we divided the annotated data into training, development, and test splits consisting of 70% training, 10%, and 20% of the data respectively.

---
[2] Although the original sentence is from BBC Pidgin https://www.bbc.com/pidgin/tori-51702073

[3] https://www.masakhane.io

| Language | Data Source | Train/ dev/ test | # Anno. | PER | ORG | LOC | DATE | % of Entities in Tokens | # Tokens |
|---|---|---|---|---|---|---|---|---|---|
| Amharic | DW & BBC | 1750/ 250/ 500 | 4 | 730 | 403 | 1,420 | 580 | 15.13 | 37,032 |
| Hausa | VOA Hausa | 1903/ 272/ 545 | 3 | 1,490 | 766 | 2,779 | 922 | 12.17 | 80,152 |
| Igbo | BBC Igbo | 2233/ 319/ 638 | 6 | 1,603 | 1,292 | 1,677 | 690 | 13.15 | 61,668 |
| Kinyarwanda | IGIHE news | 2110/ 301/ 604 | 2 | 1,366 | 1,038 | 2096 | 792 | 12.85 | 68,819 |
| Luganda | BUKEDDE news | 2003/ 200/ 401 | 3 | 1,868 | 838 | 943 | 574 | 14.81 | 46,615 |
| Luo | Ramogi FM news | 644/ 92/ 185 | 2 | 557 | 286 | 666 | 343 | 14.95 | 26,303 |
| Nigerian-Pidgin | BBC Pidgin | 2100/ 300/ 600 | 5 | 2,602 | 1,042 | 1,317 | 1,242 | 13.25 | 76,063 |
| Swahili | VOA Swahili | 2104/ 300/ 602 | 6 | 1,702 | 960 | 2,842 | 940 | 12.48 | 79,272 |
| Wolof | Lu Defu Waxu & Saabal | 1,871/ 267/ 536 | 2 | 731 | 245 | 836 | 206 | 6.02 | 52,872 |
| Yorùbá | GV & VON news | 2124/ 303/ 608 | 5 | 1,039 | 835 | 1,627 | 853 | 11.57 | 83,285 |

Table 3: Statistics of our datasets including their source, number of sentences in each split, number of annotators, number of entities of each label type, percentage of tokens that are named entities, and total number of tokens.

| Dataset | Token Fleiss' $\kappa$ | Entity Fleiss' $\kappa$ | Disagreement from Type |
|---|---|---|---|
| amh | 0.987 | 0.959 | 0.044 |
| hau | 0.988 | 0.962 | 0.097 |
| ibo | 0.995 | 0.983 | 0.071 |
| kin | 1.000 | 1.000 | 0.000 |
| lug | 0.997 | 0.990 | 0.023 |
| luo | 1.000 | 1.000 | 0.000 |
| pcm | 0.989 | 0.966 | 0.048 |
| swa | 1.000 | 1.000 | 0.000 |
| wol | 1.000 | 1.000 | 0.000 |
| yor | 0.990 | 0.964 | 0.079 |

Table 4: Inter-annotator agreement for our datasets calculated using Fleiss' kappa ($\kappa$) at the token and entity level. Disagreement from type refers to the proportion of all entity-level disagreements, which are due only to type mismatch.

A key objective of our annotation procedure was to create high-quality datasets by ensuring a high annotator agreement. To achieve high agreement scores, we ran collaborative workshops for each language, which allowed annotators to discuss any disagreements. ELISA provides an entity-level F1-score and also an interface for annotators to correct their mistakes, making it easy to achieve inter-annotator agreement scores between 0.96 and 1.0 for all languages.

We report inter-annotator agreement scores in Table 4 using Fleiss' Kappa (Fleiss, 1971) at both the token and entity level. The latter considers each span an annotator proposed as an entity. As a result of our workshops, all our datasets have exceptionally high inter-annotator agreement. For Kinyarwanda, Luo, Swahili, and Wolof, we report perfect inter-annotator agreement scores ($\kappa = 1$). For each of these languages, two annotators annotated each token and were instructed to discuss and resolve conflicts among themselves. The Appendix provides a detailed entity-level confusion matrix in Table 11.

## 5 Experimental Setup

### 5.1 NER baseline models

To evaluate baseline performance on our dataset, we experiment with three popular NER models: CNN-BiLSTM-CRF, multilingual BERT (mBERT), and XLM-RoBERTa (XLM-R). The latter two models are implemented using the HuggingFace transformers toolkit (Wolf et al., 2019). For each language, we train the models on the in-language training data and evaluate on its test data.

---
[4] https://cs.nyu.edu/~grishman/muc6.html

**CNN-BiLSTM-CRF** This architecture was proposed for NER by Ma and Hovy (2016). For each input sequence, we first compute the vector representation for each word by concatenating character-level encodings from a CNN and vector embeddings for each word. Following Rijhwani et al. (2020), we use randomly initialized word embeddings since we do not have high-quality pre-trained embeddings for all the languages in our dataset. Our model is implemented using the DyNet toolkit (Neubig et al., 2017).

**mBERT** We fine-tune multilingual BERT (Devlin et al., 2019) on our NER corpus by adding a linear classification layer to the pre-trained transformer model, and train it end-to-end. mBERT was trained on 104 languages including only two African languages: Swahili and Yorùbá. We use the mBERT-base cased model with 12-layer Transformer blocks consisting of 768-hidden size and 110M parameters.

**XLM-R** XLM-R (Conneau et al., 2020) was trained on 100 languages including Amharic, Hausa, and Swahili. The major differences between XLM-R and mBERT are (1) XLM-R was trained on Common Crawl while mBERT was trained on Wikipedia; (2) XLM-R is based on RoBERTa, which is trained with a masked language model (MLM) objective while mBERT was additionally trained with a next sentence prediction objective. We make use of the XLM-R base and large models for the baseline models. The XLM-R-base model consisting of 12 layers, with a hidden size of 768 and 270M parameters. On the other hand, the XLM-R-large has 24 layers, with a hidden size of 1024 and 550M parameters.

**MeanE-BiLSTM** This is a simple BiLSTM model with an additional linear classifier. For each input sequence, we first extract a sentence embedding from mBERT or XLM-R language model (LM) before passing it into the BiLSTM model. Following Reimers and Gurevych (2019), we make use of the mean of the 12-layer output embeddings of the LM (i.e *MeanE*). This has been shown to provide better sentence representations than the embedding of the [CLS] token used for fine-tuning mBERT and XLM-R.

**Language BERT** The mBERT and the XLM-R models only supports two and three languages under study respectively. One effective approach to adapt the pre-trained transformer models to new domains is "domain-adaptive fine-tuning" (Howard and Ruder, 2018; Gururangan et al., 2020)—fine-tuning on unlabeled data in the new domain, which also works very well when adapting to a new language (Pfeiffer et al., 2020a; Alabi et al., 2020). For each of the African languages, we performed *language-adaptive fine-tuning* on available unlabeled corpora mostly from JW300 (Agić and Vulić, 2019), indigenous news sources and XLM-R Common Crawl corpora (Conneau et al., 2020). The appendix provides the details of the unlabeled corpora in Table 10. This approach is quite useful for languages whose scripts are not supported by the multilingual transformer models like Amharic where we replace the vocabulary of mBERT by an Amharic vocabulary before we perform language-adaptive fine-tuning, similar to Alabi et al. (2020).

### 5.2 Improving the Baseline Models

In this section, we consider techniques to improve the baseline models such as utilizing gazetteers, transfer learning from other domains and languages, and aggregating NER datasets by regions. For these experiments, we focus on the PER, ORG, and LOC categories, because the gazetteers from Wikipedia do not contain DATE entities and some source domains and languages that we transfer from do not have the DATE annotation. We apply these modifications to the XLM-R model because it generally outperforms mBERT in our experiments (see Section 6).

#### 5.2.1 Gazetteers for NER

Gazetteers are lists of named entities collected from manually crafted resources such as GeoNames or Wikipedia. Before the widespread adoption of neural networks, NER methods used gazetteers-based features to improve performance (Ratinov and Roth, 2009). These features are created for each $n$-gram in the dataset and are typically binary-valued, indicating whether that $n$-gram is present in the gazetteer.

Recently, Rijhwani et al. (2020) showed that augmenting the neural CNN-BiLSTM-CRF model with gazetteer features can improve NER performance for low-resource languages. We conduct similar experiments on the languages in our dataset, using entity lists from Wikipedia as gazetteers. For Luo and Nigerian-Pidgin, which do not have their own Wikipedia, we use entity lists from English Wikipedia.

### 5.2.2 Transfer Learning

Here, we focus on cross-domain transfer from Wikipedia to the news domain, and cross-lingual transfer from English and Swahili NER datasets to the other languages in our dataset.

**Domain Adaptation from WikiAnn** We make use of the WikiAnn corpus (Pan et al., 2017), which is available for five of the languages in our dataset: Amharic, Igbo, Kinyarwanda, Swahili and Yorùbá. For each language, the corpus contains 100 sentences in each of the training, development and test splits except for Swahili, which contains 1K sentences in each split. For each language, we train on the corresponding WikiAnn training set and either zero-shot transfer to our respective test set or additionally fine-tune on our training data.

**Cross-lingual transfer** For training the cross-lingual transfer models, we use the CoNLL-2003[5] NER dataset in English with over 14K training sentences and our annotated corpus. The reason for CoNLL-2003 is because it is in the same news domain as our annotated corpus. We also make use of the languages that are supported by the XLM-R model and are widely spoken in East and West Africa like Swahili and Hausa. The English corpus has been shown to transfer very well to low resource languages (Hedderich et al., 2020; Lauscher et al., 2020). We first train on either the English CoNLL-2003 data or our training data in Swahili, Hausa, or Nigerian-Pidgin before testing on the target African languages.

### 5.3 Aggregating Languages by Regions

As previously illustrated in Table 2, several entities have the same form in different languages while some entities may be more common in the region where the language is spoken. To study the performance of NER models across geographical areas, we combine languages based on the region of Africa that they are spoken in (see Table 1): (1) East region with Kinyarwanda, Luganda, Luo, and Swahili; (2) West Region with Hausa, Igbo, Nigerian-Pidgin, Wolof, and Yorùbá languages, (3) East and West regions—all languages except Amharic because of its distinct writing system.

---

[5]We also tried OntoNotes 5.0 by combining FAC & ORG as "ORG" and GPE & LOC as "LOC" and others as "O" except "PER", but it gave lower performance in zero-shot transfer (19.38 F1) while CoNLL-2003 gave 37.15 F1.

## 6 Results

### 6.1 Baseline Models

Table 5 gives the F1-score obtained by CNN-BiLSTM-CRF, mBERT and XLM-R models on the test sets of the ten African languages when training on our in-language data. We additionally indicate whether the language is supported by the pre-trained language models (✓). The percentage of entities that are of out-of-vocabulary (OOV; entities in the test set that are not present in the training set) is also reported alongside results of the baseline models. In general, the datasets with greater numbers of OOV entities have lower performance with the CNN-BiLSTM-CRF model, while those with lower OOV rates (Hausa, Igbo, Swahili) have higher performance. We find that the CNN-BiLSTM-CRF model performs worse than fine-tuning mBERT and XLM-R models end-to-end (FTune). We expect performance to be better (e.g., for Amharic and Nigerian-Pidgin with over 18 F1 point difference) when using pre-trained word embeddings for the initialization of the BiLSTM model rather than random initialization (we leave this for future work as discussed in Section 7).

Interestingly, the pre-trained language models (PLMs) have reasonable performance even on languages they were not trained on such as Igbo, Kinyarwanda, Luganda, Luo, and Wolof. However, languages supported by the PLM tend to have better performance overall. We observe that fine-tuned XLM-R-base models have significantly better performance on five languages; two of the languages (Amharic and Swahili) are supported by the pre-trained XLM-R. Similarly, fine-tuning mBERT has better performance for Yorùbá since the language is part of the PLM's training corpus. Although mBERT is trained on Swahili, XLM-R-base shows better performance. This observation is consistent with Hu et al. (2020) and could be because XLM-R is trained on more Swahili text (Common Crawl with 275M tokens) whereas mBERT is trained on a smaller corpus from Wikipedia (6M tokens[6]).

Another observation is that mBERT tends to have better performance for the non-Bantu Niger-Congo languages i.e., Igbo, Wolof, and Yorùbá, while XLM-R-base works better for Afro-

---

[6]https://github.com/mayhewsw/multilingual-data-stats

| Lang. | In mBERT? | In XLM-R? | % OOV in Test Entities | CNN-BiLSTM CRF | mBERT-base MeanE / FTune | XLM-R-base MeanE / FTune | XLM-R Large FTune | lang. BERT FTune | lang. XLM-R FTune |
|---|---|---|---|---|---|---|---|---|---|
| amh | ✗ | ✓ | 72.94 | 52.08 | 0.0 / 0.0 | 63.57 / 70.62 | 76.18 | 60.89 | **77.97** |
| hau | ✗ | ✓ | 33.40 | 83.52 | 81.49 / 86.65 | 86.06 / 89.50 | 90.54 | 91.31 | **91.47** |
| ibo | ✗ | ✗ | 46.56 | 80.02 | 76.17 / 85.19 | 73.47 / 84.78 | 84.12 | 86.75 | **87.74** |
| kin | ✗ | ✗ | 57.85 | 62.97 | 65.85 / 72.20 | 63.66 / 73.32 | 73.75 | 77.57 | **77.76** |
| lug | ✗ | ✗ | 61.12 | 74.67 | 70.38 / 80.36 | 68.15 / 79.69 | 81.57 | 83.44 | **84.70** |
| luo | ✗ | ✗ | 65.18 | 65.98 | 56.56 / 74.22 | 52.57 / 74.86 | 73.58 | **75.59** | 75.27 |
| pcm | ✗ | ✗ | 61.26 | 67.67 | 81.87 / 87.23 | 81.93 / 87.26 | 89.02 | 89.95 | **90.00** |
| swa | ✓ | ✓ | 40.97 | 78.24 | 83.08 / 86.80 | 84.33 / 87.37 | 89.36 | 89.36 | **89.46** |
| wol | ✗ | ✗ | 69.73 | 59.70 | 57.21 / 64.52 | 54.97 / 63.86 | 67.90 | **69.43** | 68.31 |
| yor | ✓ | ✗ | 65.99 | 67.44 | 74.28 / 78.97 | 67.45 / 78.26 | 78.89 | 82.58 | **83.66** |
| avg | – | – | 57.50 | 69.23 | 64.69 / 71.61 | 69.62 / 78.96 | 80.49 | 80.69 | **82.63** |
| avg (excl. amh) | – | – | 55.78 | 71.13 | 71.87 / 79.88 | 70.29 / 79.88 | 80.97 | 82.89 | **83.15** |

Table 5: NER model comparison, showing F1-score on the test sets after 50 epochs averaged over 5 runs. This result is for all 4 tags in the dataset: PER, ORG, LOC, DATE. **Bold** marks the top score (tied if within the range of SE). mBERT and XLM-R are trained in two ways (1) MeanE: mean output embeddings of the 12 LM layers are used to initialize BiLSTM + Linear classifier, and (2) FTune: LM fine-tuned end-to-end with a linear classifier. Lang. BERT & Lang XLM-R (base) are models fine-tuned after language adaptive fine-tuning.

| Method | amh | hau | ibo | kin | lug | luo | pcm | swa | wol | yor | avg |
|---|---|---|---|---|---|---|---|---|---|---|---|
| CNN-BiLSTM-CRF | 50.31 | 84.64 | 81.25 | 60.32 | 75.66 | 68.93 | 62.60 | 77.83 | 61.84 | 66.48 | 68.99 |
| + Gazetteers | 49.51 | **85.02** | 80.40 | **64.54** | 73.85 | 65.44 | **66.54** | **80.16** | **62.44** | 65.49 | **69.34** |

Table 6: Improving NER models using Gazetteers. The result is only for 3 Tags: PER, ORG & LOC. Models trained for 50 epochs. Result is an average over 5 runs.

Asiatic languages (i.e., Amharic and Hausa), Nilo-Saharan (i.e., Luo) and Bantu languages like Kinyarwanda and Swahili. We also note that the writing script is one of the primary factors influencing the transfer of knowledge in PLMs with regard to the languages they were not trained on. For example, mBERT achieves an F1-score of 0.0 on Amharic because it has not encountered the script during pre-training. In general, we find the fine-tuned XLM-R-large (with 550M parameters) to be better than XLM-R-base (with 270M parameters) and mBERT (with 110 parameters) in almost all languages. However, mBERT models perform slightly better for Igbo, Luo, and Yorùbá despite having fewer parameters.

We further analyze the transfer abilities of mBERT and XLM-R by extracting sentence embeddings from the LMs to train a BiLSTM model (*MeanE-BiLSTM*) instead of fine-tuning them end-to-end. Table 5 shows that languages that are not supported by mBERT or XLM-R generally perform worse than CNN-BiLSTM-CRF model (despite being randomly initialized) except for *kin*. Also, sentence embeddings extracted from mBERT often lead to better performance than XLM-R for languages they both do not support (like *ibo*, *kin*, *lug*, *luo*, and *wol*).

Lastly, we train NER models using *language BERT* models that have been adapted to each of the African languages via language-specific fine-tuning on unlabeled text. In all cases, fine-tuning language BERT and language XLM-R models achieves a 1 − 7% improvement in F1-score over fine-tuning mBERT-base and XLM-R-base respectively. This approach is still effective for small sized pre-training corpora provided they are of good quality. For example, the Wolof monolingual corpus, which contains less than 50K sentences (see Table 10 in the Appendix) still improves performance by over 4% F1. Further, we obtain over 60% improvement in performance for Amharic BERT because mBERT does not recognize the Amharic script.

### 6.2 Evaluation of Gazetteer Features

Table 6 shows the performance of the CNN-BiLSTM-CRF model with the addition of gazetteer features as described in Section 5.2.1.

| Method | amh | hau | ibo | kin | lug | luo | pcm | swa | wol | yor | avg |
|---|---|---|---|---|---|---|---|---|---|---|---|
| XLM-R-base | 69.71 | 91.03 | 86.16 | 73.76 | 80.51 | 75.81 | 86.87 | **88.65** | 69.56 | 78.05 | 77.30 |
| WikiAnn zero-shot | 27.68 | – | 21.90 | 9.56 | – | – | – | 36.91 | – | 10.42 | – |
| eng-CoNLL zero-shot | – | 67.52 | 47.71 | 38.17 | 39.45 | 34.19 | 67.27 | 76.40 | 24.33 | 39.04 | 37.15 |
| pcm zero-shot | – | 63.71 | 42.69 | 40.99 | 43.50 | 33.12 | – | 72.84 | 25.37 | 35.16 | 36.81 |
| swa zero-shot | – | 85.35* | 55.37 | 58.44 | 57.65* | 42.88* | 72.87* | – | 41.70 | 57.87* | 52.32 |
| hau zero-shot | – | – | 58.41* | 59.10* | 59.78 | 42.81 | 70.74 | 83.19* | 42.81* | 55.97 | 53.14* |
| WikiAnn + finetune | **70.92** | – | 85.24 | 72.84 | – | – | – | 87.90 | – | 76.78 | – |
| eng-CoNLL + finetune | – | 89.73 | 85.10 | 71.55 | 77.34 | 73.92 | 84.05 | 87.59 | 68.11 | 75.77 | 75.30 |
| pcm + finetune | – | 90.78 | 86.42 | 71.69 | 79.72 | 75.56 | – | 87.62 | 67.21 | 78.29 | 76.48 |
| swa + finetune | – | 91.50 | 87.11 | 74.84 | 80.21 | 74.49 | 86.74 | – | 68.47 | **80.68** | 77.63 |
| hau + finetune | – | – | 86.84 | 74.22 | 80.56 | 75.55 | 88.03 | 87.92 | **70.20** | 79.44 | 77.80 |
| combined East Langs. | – | – | – | 75.65 | 81.10 | 77.56 | – | 88.15 | – | – | – |
| combined West Langs. | – | 90.88 | 87.06 | – | – | – | 87.21 | – | 69.70 | **80.68** | – |
| combined 9 Langs. | – | **91.64** | **87.94** | 75.46 | **81.29** | **78.12** | **88.12** | 88.10 | 69.84 | 80.59 | **78.87** |

Table 7: Transfer Learning Result (i.e. F1-score). 3 Tags: PER, ORG & LOC. WikiAnn, eng-CoNLL, and the annotated datasets are trained for 50 epochs. Fine-tuning is only for 10 epochs. Results are averaged over 5 runs and the total average (avg) is computed over ibo, kin, lug, luo, wol, and yor languages. The overall highest F1-score is in **bold**, and the best F1-score in zero-shot settings is indicated with an asterisk (*).

| Source Language | PER | ORG | LOC |
|---|---|---|---|
| eng-CoNLL | 36.17 | 27.00 | 50.50 |
| pcm | 21.50 | 65.33 | 68.17 |
| swa | 55.00 | 69.67 | 46.00 |
| hau | 52.67 | 57.50 | 48.50 |

Table 8: Average per-named entity F1-score for the zero-shot NER using the XLM-R model. The average is computed over ibo, kin, lug, luo, wol, yor languages.

On average, the model that uses gazetteer features performs better than the baseline. In general, languages with larger gazetteers, such as Swahili (16K entities in the gazetteer) and Nigerian-Pidgin (for which we use an English gazetteer with 2M entities), have more improvement in performance than those with fewer gazetteer entries, such as Amharic and Luganda (2K and 500 gazetteer entities respectively). This indicates that having high-coverage gazetteers is important for the model to take advantage of the gazetteer features.

### 6.3 Transfer Learning Experiments

Table 7 shows the result for the different transfer learning approaches, which we discuss individually in the following sections. We make use of XLM-R-base model for all the experiments in this sub-section because the performance difference if we use XLM-R-large is small (<2%) as shown in Table 5 and because it is faster to train.

#### 6.3.1 Cross-domain Transfer

We evaluate cross-domain transfer from Wikipedia to the news domain for the five languages that are available in the WikiAnn (Pan et al., 2017) dataset. In the zero-shot setting, the NER F1-score is low: less than 40 F1-score for all languages, with Kinyarwanda and Yorùbá having less than 10 F1-score. This is likely due to the number of training sentences present in WikiAnn: there are only 100 sentences in the datasets of Amharic, Igbo, Kinyarwanda and Yorùbá. Although the Swahili corpus has 1,000 sentences, the 35 F1-score shows that transfer is not very effective. In general, cross-domain transfer is a challenging problem, and is even harder when the number of training examples from the source domain is small. Fine-tuning on the in-domain news NER data does not improve over the baseline (XLM-R-base).

#### 6.3.2 Cross-Lingual Transfer

**Zero-shot** In the zero-shot setting we evaluated NER models trained on the English eng-*CoNLL03* dataset, and on the Nigerian-Pidgin (pcm), Swahili (swa), and Hausa (hau) annotated corpus. We excluded the MISC entity in the eng-*CoNLL03* corpus because it is absent in our target datasets. Table 7 shows the result for the (zero-shot) transfer performance. We observe that the closer the source and target languages are geographically, the bet-ter the performance. The `pcm` model (trained on only 2K sentences) obtains similar transfer performance as the `eng`-*CoNLL03* model (trained on 14K sentences). `swa` performs better than `pcm` and `eng`-*CoNLL03* with an improvement of over 14 F1 on average. We found that, on average, transferring from Hausa provided the best F1, with an improvement of over 16% and 1% compared to using the `eng`-*CoNLL* and `swa` data respectively. Per-entity analysis in Table 8 shows that the largest improvements are obtained for `ORG`. The `pcm` data was more effective in transferring to `LOC` and `ORG`, while `swa` and `hau` performed better when transferring to `PER`. In general, zero-shot transfer is most effective when transferring from Hausa and Swahili.

| Language | CNN-BiLSTM | | | | | mBERT-base | | | | | XLM-R-base | | | | |
|---|---|---|---|---|---|---|---|---|---|---|---|---|---|---|---|
| | all | 0-freq | 0-freq Δ | long | long Δ | all | 0-freq | 0-freq Δ | long | long Δ | all | 0-freq | 0-freq Δ | long | long Δ |
| amh | 52.89 | 40.98 | -11.91 | 45.16 | -7.73 | – | – | – | – | – | 70.96 | 68.91 | -2.05 | 64.86 | -6.10 |
| hau | 83.70 | 78.52 | -5.18 | 66.21 | -17.49 | 87.34 | 79.41 | -7.93 | 67.67 | -19.67 | 89.44 | 85.48 | -3.96 | 76.06 | -13.38 |
| ibo | 78.48 | 70.57 | -7.91 | 53.93 | -24.55 | 85.11 | 78.41 | -6.70 | 60.46 | -24.65 | 84.51 | 77.42 | -7.09 | 59.52 | -24.99 |
| kin | 64.61 | 55.89 | -8.72 | 40.00 | -24.61 | 70.98 | 65.57 | -5.41 | 55.39 | -15.59 | 73.93 | 66.54 | -7.39 | 54.96 | -18.97 |
| lug | 74.31 | 67.99 | -6.32 | 58.33 | -15.98 | 80.56 | 76.27 | -4.29 | 65.67 | -14.89 | 80.71 | 73.54 | -7.17 | 63.77 | -16.94 |
| luo | 66.42 | 58.93 | -7.49 | 54.17 | -12.25 | 72.65 | 72.85 | 0.20 | 66.67 | -5.98 | 75.14 | 72.34 | -2.80 | 69.39 | -5.75 |
| pcm | 66.43 | 59.73 | -6.70 | 47.80 | -18.63 | 87.78 | 82.40 | -5.38 | 77.12 | -10.66 | 87.39 | 83.65 | -3.74 | 74.67 | -12.72 |
| swa | 79.26 | 64.74 | -14.52 | 44.78 | -34.48 | 86.37 | 78.77 | -7.60 | 45.55 | -40.82 | 87.55 | 80.91 | -6.64 | 53.93 | -33.62 |
| wol | 60.43 | 49.03 | -11.40 | 26.92 | -33.51 | 66.10 | 59.54 | -6.56 | 19.05 | -47.05 | 64.38 | 57.21 | -7.17 | 38.89 | -25.49 |
| yor | 67.07 | 56.33 | -10.74 | 64.52 | -2.55 | 78.64 | 73.41 | -5.23 | 74.34 | -4.30 | 77.58 | 72.01 | -5.57 | 76.14 | -1.44 |
| avg (excl. amh) | 69.36 | 60.27 | -9.09 | 50.18 | -19.18 | 79.50 | 74.07 | -5.43 | 59.10 | -20.40 | 79.15 | 73.80 | -5.36 | 63.22 | -15.94 |

Table 9: F1 score for two varieties of hard-to-identify entities: zero-frequency entities that do not appear in the training corpus, and longer entities of four or more words.

**Fine-tuning** We use the target language corpus to fine-tune the NER models previously trained on `eng`-*CoNLL*, `pcm`, and `swa`. On average, there is only a small improvement when compared to the XLM-R base model. In particular, we see significant improvement for Hausa, Igbo, Kinyarwanda, Nigerian-Pidgin, Wolof, and Yorùbá using either `swa` or `hau` as the source NER model.

### 6.4 Regional Influence on NER

We evaluate whether combining different language training datasets by region affects the performance for individual languages. Table 7 shows that all languages spoken in West Africa (`ibo`, `wol`, `pcm`, `yor`) except `hau` have slightly better performance (0.1–2.6 F1) when we train on their combined training data. However, for the East-African languages, the F1 score only improved (0.8–2.3 F1) for three languages (`kin`, `lug`, `luo`). Training the NER model on all nine languages leads to better performance on all languages except Swahili. On average over six languages (`ibo`, `kin`, `lug`, `luo`, `wol`, `yor`), the performance improves by 1.6 F1.

### 6.5 Error analysis

Finally, to better understand the types of entities that were successfully identified and those that were missed, we performed fine-grained analysis of our baseline methods mBERT and XLM-R using the method of Fu et al. (2020), with results shown in Table 9. Specifically, we found that across all languages, entities that were not contained in the training data (zero-frequency entities), and entities consisting of more than three words (long entities) were particularly difficult in all languages; compared to the F1 score over all entities, the scores dropped by around 5 points when evaluated on zero-frequency entities, and by around 20 points when evaluated on long entities. Future work on low-resource NER or cross-lingual representation learning may further improve on these hard cases.

## 7 Conclusion and Future Work

We address the NER task for African languages by bringing together a variety of stakeholders to create a high-quality NER dataset for ten African languages. We evaluate multiple state-of-the-art NER models and establish strong baselines. We have released one of our best models that can recognize named entities in ten African languages on HuggingFace Model Hub[7]. We also investigate cross-domain transfer with experiments on five languages with the WikiAnn dataset, along with cross-lingual transfer for low-resource NER using the English CoNLL-2003 dataset and other languages supported by XLM-R. In the future, we plan to use pretrained word embeddings such as GloVe (Pennington et al., 2014) and fastText (Bojanowski et al., 2017) instead of random initialization for the CNN-BiLSTM-CRF, increase the number of annotated sentences per language, and expand the dataset to more African languages.

---
[7] https://huggingface.co/Davlan/xlm-roberta-large-masakhaner

## Acknowledgements

We would like to thank Heng Ji and Ying Lin for providing the ELISA NER tool used for annotation. We also thank the Spoken Language Systems Chair, Dietrich Klakow at Saarland University for providing GPU resources to train the models. We thank Adhi Kuncoro and the anonymous reviewers for their useful feedback on a draft of this paper. David Adelani acknowledges the support of the EU-funded H2020 project COMPRISE under grant agreement No. 3081705. Finally, we thank Mohamed Ahmed for proof-reading the draft.

| Language | Source | Size (MB) | No. sentences |
|---|---|---|---|
| `amh` | CC-100 (Conneau et al., 2020) | 889.7MB | 3,124,760 |
| `hau` | CC-100 | 318.4MB | 3,182,277 |
| `ibo` | JW300 (Agić and Vulić, 2019), CC-100, CC-Aligned (El-Kishky et al., 2020), and IgboNLP (Ezeani et al., 2020) | 118.3MB | 1,068,263 |
| `kin` | JW300, KIRNEWS (Niyongabo et al., 2020), and BBC Gahuza | 123.4MB | 726,801 |
| `lug` | JW300, CC-100, and BUKEDDE News | 54.0MB | 506,523 |
| `luo` | JW300 | 12.8MB | 160,904 |
| `pcm` | JW300, and BBC Pidgin | 56.9MB | 207,532 |
| `swa` | CC-100 | 1,800MB | 12,664,787 |
| `wol` | OPUS (Tiedemann, 2012) (excl. CC-Aligned), Wolof Bible (MBS, 2020), and news corpora (Lu Defu Waxu, Saabal, and Wolof Online) | 3.8MB | 42,621 |
| `yor` | JW300, Yoruba Embedding Corpus (Alabi et al., 2020), MENYO-20k (Adelani et al., 2021), CC-100, CC-Aligned, and news corpora (BBC Yoruba, Asejere, and Alaroye). | 117.6MB | 910,628 |

Table 10: Monolingual Corpora, their sources, size, and number of sentences

## A Appendix

### A.1 Annotator Agreement

To shed more light on the few cases where annotators disagreed, we provide entity-level confusion matrices across all ten languages in Table 11. The most common disagreement is between organizations and locations.

|  | DATE | LOC | ORG | PER |
|---|---|---|---|---|
| **DATE** | 32,978 | - | - | - |
| **LOC** | 10 | 70,610 | - | - |
| **ORG** | 0 | 52 | 35,336 | - |
| **PER** | 2 | 48 | 12 | 64,216 |

Table 11: Entity-level confusion matrix between annotators, calculated over all ten languages.

### A.2 Model Hyper-parameters for Reproducibility

For fine-tuning mBERT and XLM-R, we used the base and large models with maximum sequence length of 164 for mBERT and 200 for XLM-R, batch size of 32, learning rate of 5e-5, and number of epochs 50. For the MeanE-BiLSTM model, the hyper-parameters are similar to fine-tuning the LM except for the learning rate that we set to be 5e-4, the BiLSTM hyper-parameters are: input dimension is 768 (since the embedding size from mBERT and XLM-R is 768) in each direction of LSTM, one hidden layer, hidden layer size of 64, and drop-out probability of 0.3 before the last linear layer. All the experiments were performed on a single GPU (Nvidia V100).

### A.3 Monolingual Corpora for Language Adaptive Fine-tuning

Table 10 shows the monolingual corpus we used for the language adaptive fine-tuning. We provide the details of the source of the data, and their sizes. For most of the languages, we make use of JW300[8] and CC-100[9]. In some cases CC-Aligned (El-Kishky et al., 2020) was used, in such a case, we removed duplicated sentences from CC-100. For fine-tuning the language model, we make use of the HuggingFace (Wolf et al., 2019) code with learning rate 5e-5. However, for the Amharic BERT, we make use of a smaller learning rate of 5e-6 since the multilingual BERT vocabulary was replaced by Amharic vocabulary, so that we can slowly adapt the mBERT LM to understand Amharic texts. All language BERT models were pre-trained for 3 epochs ("ibo", "kin","lug","luo", "pcm","swa","yor") or 10 epochs ("amh", "hau","wol") depending on their convergence. The models can be found on HuggingFace Model Hub[10].

---
[8] https://opus.nlpl.eu/
[9] http://data.statmt.org/cc-100/
[10] https://huggingface.co/Davlan